\documentclass[11pt,a4paper]{article}
\usepackage[hyperref]{eacl2021}
\usepackage{times}
\usepackage{graphicx}
\usepackage{amsmath}
\usepackage{amssymb}
\usepackage{latexsym}

\usepackage{enumitem}

\usepackage{microtype}

\aclfinalcopy %

\renewcommand{\vec}[1]{\mathbf{#1}}

\DeclareMathOperator*{\argmax}{arg\,max}

\title{Fast and Effective Biomedical Entity Linking Using a Dual Encoder}

\author{Rajarshi Bhowmik \and
  Karl Stratos \\
  Rutgers University--New Brunswick \\ Piscataway, New Jersey, USA \\
  \texttt{rajarshi.bhowmik@rutgers.edu}\\
  \texttt{karl.stratos@rutgers.edu}
  \And Gerard de Melo \\
  Hasso Plattner Institute \\ 
  University of Potsdam\\ 
  Potsdam, Germany \\
  \texttt{gdm@demelo.org}
  }

\date{}

\begin{document}
\maketitle
\begin{abstract}
    Biomedical entity linking is the task of identifying mentions of biomedical concepts in text documents and mapping them to canonical entities in a target thesaurus. Recent advancements in entity linking using BERT-based models follow a \emph{retrieve and rerank} paradigm, where the candidate entities are first selected using a retriever model, and then the retrieved candidates are ranked by a reranker model. While this paradigm produces state-of-the-art results, they are slow both at training and test time as they can process only one mention at a time. 
    To mitigate these issues, we propose a BERT-based dual encoder model that resolves multiple mentions in a document in one shot. We show that our proposed model is multiple times faster than existing BERT-based models while being competitive in accuracy for biomedical entity linking. Additionally, we modify our dual encoder model for end-to-end biomedical entity linking that performs both mention span detection and entity disambiguation and out-performs two recently proposed models. 
\end{abstract}

\section{Introduction}
Entity linking is the task of identifying mentions of named entities (or  other terms) in a text document and disambiguating them by mapping them to canonical entities (or concepts) listed in a reference knowledge graph \cite{KnowledgeGraphs2020}. This is an essential step in information extraction, and therefore has been studied extensively both in domain-specific and domain-agnostic settings. Recent state-of-the-art models
\cite{logeswaran-etal-2019-zero, BLINK:2019:Wu} attempt
to learn better representations of mentions and candidates using the rich contextual information encoded in pre-trained language models such as BERT \cite{devlin-etal-2019-bert}.
These models follow a \emph{retrieve and rerank} paradigm, which consists of two separate steps: First, the candidate entities are selected using a retrieval model. Subsequently, the retrieved candidates are ranked by a reranker model.

Although this approach has yielded strong results, owing primarily to the powerful contextual representation learning ability of BERT-based encoders, these models typically process a single mention at a time. 
Processing one mention at a time incurs a substantial overhead both during training and test time, leading to a system that is slow and impractical.

In this paper, we propose a \emph{collective entity linking} method that  processes an entire document only once, such that all entity mentions within it are linked to their respective target entities in the knowledge base in one pass.

Compared to the popular
entity linking model BLINK \cite{BLINK:2019:Wu}, our model is up to 25x faster. BLINK deploys two separately trainable models for candidate retrieval and reranking. In contrast, our method learns a single model that can perform both the retrieval and reranking steps of entity linking. Our model does not require candidate retrieval at inference time, as our dual encoder approach allows us to compare each mention to all entities in the target knowledge base, thus significantly reducing the overhead at inference time.

We evaluate our method on two particularly challenging datasets from the biomedical domain.
In recent times, there is an increased focus on information extraction from biomedical text such as biomedical academic publications, electronic health records, discharge summaries of patients, or clinical reports. Extracting named concepts from biomedical text requires domain expertise. Existing automatic extraction methods, including the methods and tools catering to the biomedical domain \cite{savova2010ctakes, soldaini2016quickumls, aronson2006metamap}, often perform poorly due to the inherent challenges of biomedical text: (1) Biomedical text typically contains substantial domain-specific jargon and abbreviations. For example, \emph{CT} could stand for \emph{Computed tomography} or \emph{Copper Toxicosis}. (2) The target concepts in the knowledge base often have very similar surface forms, making the disambiguation task difficult. For example, \emph{Pseudomonas aeruginosa} is a kind of bacteria, while \emph{Pseudomonas aeruginosa infection} is a disease. Many existing biomedical information extraction tools rely on similarities in surface forms of mentions and candidates, and thus invariably falter in more challenging cases such as these. 
Additionally, long mention spans (e.g., disease names) and the density of mentions per document make the biomedical entity linking very challenging. 

\paragraph{Contributions} 
The key contributions of our work are as follows.
\begin{itemize}[noitemsep,nolistsep]

    \item Training our collective entity disambiguation model is 3x faster than other dual encoder models with the same number of parameters that perform per-mention entity disambiguation. At inference time, our model is 3-25x faster than other comparable models. 
    
    \item At the same time, our model obtains favorable results on two biomedical datasets compared to state-of-the-art entity linking models.
    
    \item Our model can also perform end-to-end entity linking when trained with the multi-task objective of mention span detection and entity disambiguation. We show that without using any semantic type information, our model significantly out-performs two recent biomedical entity linking models -- MedType \cite{medtype2020} and SciSpacy \cite{neumann-etal-2019-scispacy} -- on two benchmark datasets.
\end{itemize}

\begin{figure*}
    \centering
    \includegraphics[width=0.9\linewidth]{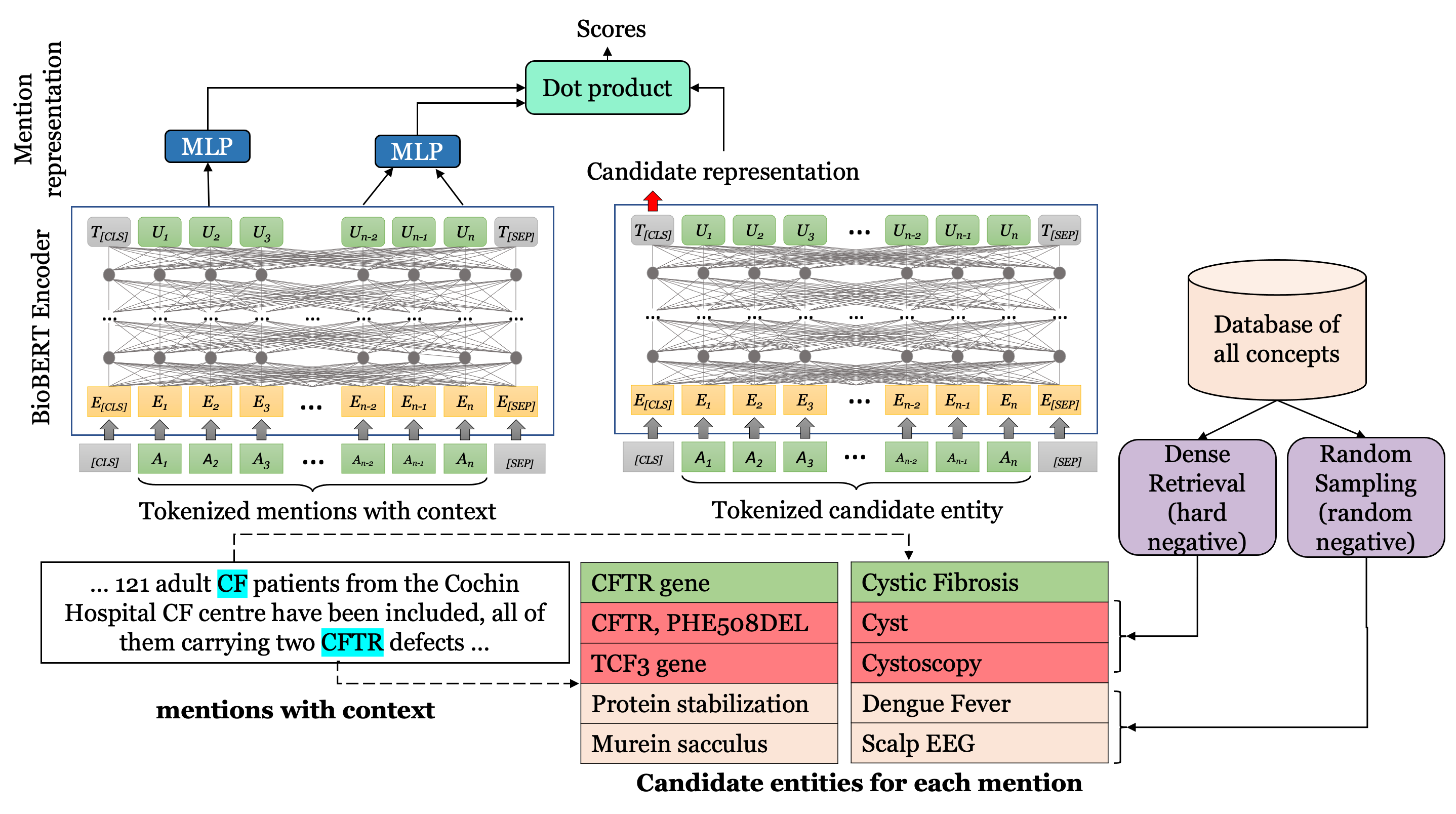}
    \caption{A schematic diagram of the Dual Encoder model for collective entity disambiguation. In this diagram, the number of mentions in a document and the number of candidate entities per mention are for illustration purpose only. The inputs to the BioBERT encoders are the tokens obtained from the BioBERT tokenizer.}
    \label{fig:model_diagram}
\end{figure*}

\section{Related Work}
\subsection{Entity Linking} 
The task of entity linking has been studied extensively in the literature. In the past, most models relied on hand-crafted features for entity disambiguation using surface forms and alias tables, which may not be available for every domain. With the advent of deep learning, contextual representation learning for mention spans has become more popular.
Recent Transformer-based models for entity linking \cite{BLINK:2019:Wu, fvry2020empirical} have achieved state-of-the-art performance on traditional benchmark datasets such as AIDA-CoNLL and TACKBP 2010.

\subsection{Biomedical Entity Linking} 
In the biomedical domain, there are many existing tools, such as TaggerOne \cite{leaman2016taggerone}, MetaMap \cite{aronson2006metamap}, cTAKES \cite{savova2010ctakes}, QuickUMLS \cite{soldaini2016quickumls}, among others, for normalizing mentions of biomedical concepts to a biomedical thesaurus. Most of these methods rely on feature-based approaches. Recently, \newcite{LATTE:Zhu:2019}
proposed a model that utilizes the latent semantic information of mentions and entities to perform entity linking. Other recent models such as \newcite{xu-etal-2020-generate} and \newcite{medtype2020} also leverage semantic type information for improved entity disambiguation. Our work is different from these approaches, as our model does not use semantic type information, since such information may not always be available. 
Recent studies such as \newcite{xu-etal-2020-generate} and \newcite{ji2020bert} deploy a BERT-based retrieve and re-rank model. In contrast, our model does not rely on a separate re-ranker model, which significantly improves its efficiency.

\subsection{End-to-End Entity Linking}
End-to-end entity linking refers to the task of predicting mention spans and the corresponding target entities jointly using a single model. Traditionally, span detection and entity disambiguation tasks were done in a pipelined approach, making these approaches susceptible to error propagation. To alleviate this issue, \newcite{kolitsas-etal-2018-end} proposed a neural end-to-end model that performs the dual tasks of mention span detection and entity disambiguation. However, for span detection and disambiguation, their method relies on an empirical probabilistic entity mapping $p(e|m)$ to select a candidate set $C(m)$ for each mention $m$. Such mention--entity prior $p(e|m)$ is not available in every domain, especially in the biomedical domain that we consider in this paper. In contrast, our method does not rely on any extrinsic sources of information. Recently, \newcite{furrer2020parallel} proposed a parallel sequence tagging model that treats both span detection and entity disambiguation as sequence tagging tasks. However, one practical disadvantage of their model is the large number of tag labels when the target knowledge base contains thousands of entities. In contrast, our dual encoder model can effectively link mentions to a knowledge base with large number of entities.
\section{Model}

Given a document $d = [x_1^d, \dots, x_T^d]$ of $T$ tokens with $N$ mentions $\{m_1, \dots , m_N\}$ and a set of $M$ entities $\{e_1, \dots, e_M\}$ in a target knowledge base or thesaurus $\mathcal{E}$, the task of collective entity disambiguation consists in mapping each entity mention $m_k$ in the document to a target entity $t_k \in \mathcal{E}$ in one shot. 
Each mention in the document $d$ may span over one or multiple tokens,  denoted by pairs $(i, j)$ of start and end index positions such that $m_k = [x_i^d, \dots, x_j^d]$. 

\subsection{Encoding Mentions and Candidates}

Our model consists of two BERT-based encoders. The \emph{mention encoder} is responsible for learning representations of contextual mentions and the \emph{candidate encoder} learns representations for the candidate entities. A schematic diagram of the model is presented in Figure \ref{fig:model_diagram}. Following the BERT model, the input sequences to these encoders start and end with the special tokens \texttt{[CLS]} and \texttt{[SEP]}, respectively.

\paragraph{Mention Encoder}
Given an input text document $[x_1^d, \dots, x_T^d]$ of $T$ tokens with $M$ mentions, the output of the final layer of the encoder, denoted by $[\vec{h_1}, \dots, \vec{h_T}]$, is a contextualized representation of the input tokens.  For each mention span $(i, j)$, we concatenate the first and the last tokens of the span and pass it through a linear layer to obtain the representations for each of the mentions. Formally, the representation of mention $m_k$ is given as 
\begin{equation}
    \vec{u_k^m} = \vec{W}[\vec{h_i};  \vec{h_j}] + \vec{b}.
\end{equation}
Since the encoder module deploys a self-attention mechanism, every mention inherently captures contextual information from the other mentions in the document. 

\paragraph{Candidate Encoder}
Given an input candidate entity $e = [y_1^e, \dots, y_T^e]$ of $T$ tokens, the output of the final layer corresponding to the \texttt{[CLS]} token yields the representation for the candidate entity. We denote the representation of entity $e$ as $\vec{v^e}$. As shown in Figure \ref{fig:model_diagram}, we use the UMLS concept name of each candidate entity as the input to the candidate encoder. 

\subsection{Candidate Selection}
\label{candidate_selection}
\paragraph{Candidate Retrieval}
Since the entity disambiguation task is formulated as a learning to rank problem, we need to retrieve negative candidate entities for ranking during training. To this end, we randomly sample a set of negative candidates from the pool of all entities in the knowledge base. Additionally, we adopt the \emph{hard negative mining} strategy used by \newcite{gillick-etal-2019-learning} to retrieve negative candidates by performing nearest neighbor search using the dense representations of mentions and candidates described above. The hard negative candidates are the entities that are more similar to the mention than the gold target entity.

\paragraph{Candidate Scoring}
The retrieved set of candidate entities $\mathcal{C}_k = \{c_1^k, \dots, c_l^k\}$ for each mention $m_k$ are scored using a dot product between the mention representation $\vec{u_k^m}$ and each candidate representation $\vec{v^c}$. Formally, for each $c \in \mathcal{C}_k$
\begin{equation}
\psi(m_k, c) = (\vec{u_k^m})^\intercal \vec{v^c} 
\end{equation}

\subsection{Training and Inference}
\label{training_and_inference}
\paragraph{Loss Function and Training}
We train our model using the cross-entropy loss function to maximize the score of the gold target entities.

\paragraph{Inference}
During inference, we do not require candidate retrieval per mention. The representations of all entities in the knowledge base $\mathcal{E}$ can be pre-computed and cached. The inference task is thus reduced to finding the maximum dot product between each mention representation and all entity representations.

\begin{equation}
    \hat{t}_k = \argmax_{e \in \mathcal{E}} \{ (\vec{u_k^m})^\intercal \vec{v^{e}} \} 
\end{equation}

\subsection{End-to-End Entity Linking}
Many of the state-of-the-art entity disambiguation models assume that gold mention spans are available during test time and thus have limited applicability in real-world entity linking tasks, where such gold mentions are typically not available. To avoid this, recent works \cite{kolitsas-etal-2018-end, fvry2020empirical,li-etal-2020-efficient} have investigated end-to-end entity linking, where a model needs to perform both mention span detection and entity disambiguation. 
\paragraph{Mention Span Detection}
We experiment with two different methods for mention span detection with different computational complexity. In our first method, following \newcite{fvry2020empirical}, we use a simple \texttt{BIO} tagging scheme to identify the mention spans. Every token in the input text is annotated with one of these three tags. Under this tagging scheme, any contiguous segment of tokens starting with a \texttt{B} tag and followed by \texttt{I} tags is treated as a mention. Although this method is computationally efficient ($\mathcal{O}(T)$), our empirical results suggest that it is not as effective as the following.

Following the recent work of \newcite{kolitsas-etal-2018-end} and \newcite{li-etal-2020-efficient}, our mention span detection method enumerates all possible spans in the input text document as potential mentions. However, enumerating all possible spans in a document of length $T$ is prohibitively large ($\mathcal{O}(T^2)$) and computationally expensive. Therefore, we constrain the maximum length of a mention span to $L \ll T$. 

We calculate the probability of each candidate mention span $(i, j)$ as follows.
\begin{equation}
    p(m|(i, j)) = \sigma \large(\vec{w}_{\mathbf{s}}^\intercal \vec{h_i} + \vec{w}_\mathbf{e}^\intercal \vec{h_j} + \sum_{q=i}^{j} \vec{w}_\mathbf{m}^\intercal \vec{h_q} \large)
\end{equation}
where $\vec{w_s}$, $\vec{w_e}$, and $\vec{w_m}$ are trainable parameters and $\sigma(x) = \frac{1}{1 + e^{-x}}$.

\paragraph{Entity Disambiguation} We represent each mention $(i, j)$ by mean pooling the final layer of the encoder, i.e., $\vec{u^m_{(i,j)}} = \frac{1}{j - i + 1} \sum_{q=i}^{j} \vec{h_q}$. During training, we perform candidate selection as described in Section \ref{candidate_selection}.

We jointly train the model by minimizing the sum of mention detection loss and entity disambiguation loss. We use a binary cross-entropy loss for mention detection with the gold mention spans as positive and other candidate mention spans as negative samples. For entity disambiguation, we use the cross-entropy loss to minimize the negative log likelihood of the gold target entity given a gold mention span.

During inference, we choose only the candidate mentions with $p(m|(i, j)) > \gamma$ as the predicted mention spans. Then, as mentioned in Section \ref{training_and_inference}, we determine the maximum dot product between the mention representations and all candidate entity representations to predict the entity for each predicted mention during inference.

\section{Evaluation}
\subsection{Datasets}
Our experiments are conducted on two challenging datasets from the biomedical domain -- MedMentions \cite{Mohan2019MedMentionsAL} and the BioCreative V Chemical Disease Relation (BC5CDR) dataset \cite{li2016biocreative}. In the following, we provide some details of these two datasets, while basic statistics are given in Table \ref{tab:datasets}.

\begin{table*}[h!]
    \centering
    \begin{tabular}{l|r|c|r|r}
        Datasets & Mentions & Mentions/Doc & Unique Concepts & Types\\
        \hline
        MedMentions & 352,496 & 80 & 34,724 & 128 \\
        BC5CDR &  28,559 & 19 & 9,149 & 2\\
        \hline
    \end{tabular}
    \caption{Details of the datasets used for evaluation.}
    \label{tab:datasets}
\end{table*}

MedMentions is a large-scale biomedical corpus annotated with UMLS concepts. It consists of a total of $4,392$ English language abstracts published on PubMed\textsuperscript{®}. The dataset has $352,496$ mentions, and each mention is associated with a single UMLS Concept Unique Identifier (CUI) and one or more semantic types identified by a Type Unique Identifier (TUI). The concepts belong to $128$ different semantic types. MedMentions also provides a 60\% -- 20\% -- 20\% random partitioning of the corpus into training, development, and test sets. Note that 12\% of the concepts in the test dataset do not occur in the training or development sets. For this dataset, our target KB consists of the concepts that are linked to at least one mention in the MedMentions dataset. 

The BC5CDR corpus consists of $1,500$ English language PubMed\textsuperscript{®} articles with $4,409$ annotated chemicals and $5,818$ diseases, which are equally partitioned into training, development, and test sets. Each entity annotation includes both the mention text spans and normalized concept identifiers, using MeSH as the target vocabulary. Apart from entity linking annotations, this dataset also provides $3,116$ chemical--disease relations. However, identifying relations between mentions is beyond the scope of our study on entity linking and hence, we ignore these annotations. 

\subsection{Baselines}
We compare our model against some of the recent state-of-the-art entity linking models from both the biomedical and non-biomedical domains. 
In the biomedical domain, LATTE \cite{LATTE:Zhu:2019} showed state-of-the-art results on the MedMentions dataset. However, we find that LATTE adds the gold target entity to the set of candidates retrieved by the BM25 retrieval method during both training and inference.

The Cross Encoder model proposed by \newcite{logeswaran-etal-2019-zero}, which follows a \emph{retrieve and rerank} paradigm, has been successfully adopted in the biomedical domain by \newcite{xu-etal-2020-generate} and \newcite{ji2020bert}. This model uses a single encoder. The input to this encoder is a concatenation of a mention with context and a candidate entity with a \texttt{[SEP]} token in between. This allows cross-attention between mentions and candidate entities. We use our own implementation of the model by \newcite{logeswaran-etal-2019-zero} for comparison. 

We also compare with BLINK \cite{BLINK:2019:Wu}, a state-of-the-art entity linking model that uses dense retrieval using dual encoders for candidate generation, followed by a cross-encoder for reranking. 

Additionally, we use the dual encoder model that processes each mention independently as a baseline. In principle, this baseline is similar to the retriever model of \newcite{BLINK:2019:Wu} and \newcite{gillick-etal-2019-learning}. 

For the task of end-to-end entity disambiguation, we compare our models with two recent state-of-the-art models -- SciSpacy \cite{neumann-etal-2019-scispacy} and MedType \cite{medtype2020}. SciSpacy uses overlapping character N-grams for mention span detection and entity disambiguation. MedType improves the results of SciSpacy by using a better candidate retrieval method that exploits the semantic type information of the candidate entities.

\begin{table*}[t]
    \centering
    \begin{tabular}{l|c|c|c|c|c|c}
    \hline
    & \multicolumn{2}{c}{Candidate retrieval method} & \multicolumn{2}{c}{Unnormalized} & \multicolumn{2}{c}{Normalized} \\
    \hline
    Model &  Training &
    Test & P@1 & MAP & P@1 & MAP \\
    \hline
    $\dagger$ LATTE & BM25 & BM25 & - & - & 88.5 & 92.8 \\
    $\dagger$ Cross Encoder & BM25 & BM25 & - & - & 91.6 & 95.1 \\
    \hline
    Cross Encoder & BM25 & BM25 & 53.8 & 56.2 & 90.4 & 94.4 \\
    Dual Encoder ($1$ mention) & DR (random) & all entities & 54.1 & 64.8 & N/A & N/A \\
    Dual Encoder ($1$ mention) & DR (random + hard) & all entities & 62.9 & 69.7 & N/A & N/A \\
    BLINK & DR (random + hard) & DR (hard) & 68.1 & 73.0 & 84.7 & 90.8 \\
    \hline
    Dual Encoder (collective) & DR (random) & all entities & 58.2 & 68.5 & N/A & N/A \\
    Dual Encoder (collective) & DR (random + hard) & all entities & \textbf{68.4} & \textbf{75.6} & N/A & N/A \\
    \hline
    \end{tabular}
    \caption{Precision@1 and Mean Average Precision (MAP) for the entity disambiguation task on the MedMentions dataset when the gold mention spans are known. $\dagger$ LATTE results are copied from the original paper and always incorporate gold entities as candidates (thus recall is always 100\%). $\dagger$ Cross Encoder shows results in this setting as a reference point. Models without $\dagger$ do not add gold entities to the candidate set.
    'N/A' stands for 'Not Applicable'. 'DR' stands for dense retrieval.}
    \label{tab:medmention-results}
\end{table*}

\begin{table*}[h]
    \centering
    \begin{tabular}{l|c|c|c|c|c|c}
    \hline
    & \multicolumn{2}{c}{Candidate retrieval method} & \multicolumn{2}{c}{Unnormalized} & \multicolumn{2}{c}{Normalized} \\
    \hline
    Model &  Training &
    Test & P@1 & MAP & P@1 & MAP \\
    \hline
    Cross Encoder & BM25 & BM25 & 72.1 & 73.1 & 96.8 & 98.1 \\
    Dual Encoder ($1$ mention) & DR (random) & all entities & 76.3 & 82.4 & N/A & N/A \\
    Dual Encoder ($1$ mention) & DR (random + hard) & all entities & \textbf{84.8} & \textbf{87.7} & N/A & N/A \\
    BLINK & DR (random + hard) & DR (hard) & 74.7 & 75.6 & 97.2 & 98.4 \\
    \hline
    Dual Encoder (collective) & DR (random) & all entities & 69.0 & 77.2 & N/A & N/A \\
    Dual Encoder (collective) & DR (random + hard) & all entities & 80.7 & 85.1 & N/A & N/A \\
    \hline
    \end{tabular}
    \caption{Precision@1 and Mean Average Precision (MAP) for the entity disambiguation task on the BC5CDR dataset when the gold mention spans are known. 'N/A' stands for 'Not Applicable'. 'DR' stands for dense retrieval.}
    \label{tab:bc5cdr-results}
\end{table*}

\subsection{Experimental Details}
In this section, we provide details pertaining to the experiments for the purpose of reproducibility. We also make the code publicly available \footnote{https://github.com/kingsaint/BioMedical-EL}.

\paragraph{Domain-Adaptive Pretraining} Recent studies \cite{logeswaran-etal-2019-zero, fvry2020empirical, BLINK:2019:Wu} have shown that pre-training BERT on the target domain provides additional performance gains for entity linking. Following this finding, we adopt BioBERT \cite{BioBERT:Lee_2019} as our domain-specific pretrained model. BioBERT is intitialzed with the parameters of the original BERT model, and  further pretrained on PubMed abstracts to adapt to biomedical NLP tasks. %

\paragraph{Data Wrangling} In theory, our collective entity disambiguation model is capable of processing documents of arbitrary length. However, there are practical constraints. First, the GPU memory limit enforces an upper bound on the number of mentions that can be processed together, and secondly, BERT stipulates the maximum length of the input sequence to be  $512$ tokens. To circumvent these constraints, we segment each document so that each chunk contains a maximum of $8$ mentions or a maximum of $512$ tokens (whichever happens earlier). After this data wrangling process, the $4,392$ original documents in the MedMentions dataset are split into $44,983$ segmented documents. Note that during inference our model can process more than $8$ mentions. However, without loss of generality, we assumed the same segmentation method during inference.
We postulate that with more GPU memory and longer context \cite{Beltagy2020Longformer}, our collective entity disambiguation model will be able to process documents of arbitrary length without segmentation during training and inference. 

For the other baselines, we process each mention along with its contexts independently. We found that a context window of $128$ characters surrounding each mention suffices for these models. We also experimented with longer contexts and observed that the performance of the models deteriorates. 

\paragraph{Hyperparameters}
To encode mentions, we use a context window of up to $128$ tokens for the single-mention Dual Encoder. The candidate entities are tokenized to a maximal length of $128$ tokens across all Dual Encoder models. In the Cross Encoder and BLINK models, where candidate tokens are appended to the context tokens, we use a maximum of $256$ tokens. For Collective Dual Encoder models, the mention encoder can encode a tokenized document of maximum length $512$. For all our experiments, we use AdamW stochastic optimization and a linear scheduling for the learning rate of the optimizer. For the single-mention Dual Encoder, Cross Encoder and BLINK model, we find an initial learning rate of $0.00005$ to be optimal. For collective Dual Encoder models, we find an initial learning rate of $0.00001$ to be suitable for both the end-to-end and non-end-to-end settings.  The ratio of hard and random negative candidates is set to 1:1, as we choose $10$ samples from each. For each model, the hyperparameters are tuned using the validation set. For the end-to-end entity linking model, we set the maximum length of a mention span $L$ to $10$ tokens.

\subsection{Evaluation Metrics} 
Picking the correct target entity among a set of candidate entities is a learning to rank problem. Therefore, we use Precision@1 and Mean Average Precision (MAP) as our evaluation metrics when the gold mention spans are known. Since there is only one correct target entity per mention in our datasets, Precision@1 is also equivalent to the accuracy. One can consider these metrics in normalized and unnormalized settings. The normalized setting is applicable when candidate retrieval is done during inference and the target entity is present in the set of retrieved candidates. Since our model and other Dual Encoder based models do not require retrieval at test time, the normalized evaluation setting is not applicable in these cases.

\subsection{Results}
\paragraph{Entity Disambiguation}
We provide the results of our experiments for the entity disambiguation task on the MedMentions and BC5CDR datasets in Tables \ref{tab:medmention-results} and \ref{tab:bc5cdr-results}, respectively. For the MedMentions dataset, our collective dual encoder model outperforms all other models, while being extremely time efficient during training and inference. On the BC5CDR dataset, our method performs adequately as compared to other baselines. Our model compares favorably against the state-of-the-art entity linking model BLINK on both datasets. Surprisingly, for the BC5CDR dataset, BLINK is outperformed by the Dual Encoder baselines that process each mention independently, despite the fact that BLINK's input candidates are generated by this model. We conjecture that BLINK's cross encoder model for re-ranking is more susceptible to overfitting on this relatively small-scale dataset.
Our model consistently outperforms the Cross Encoder model, which reinforces the prior observations made by \newcite{BLINK:2019:Wu} that dense retrieval of candidates improves the accuracy of entity disambiguation models. Finally, comparisons with an ablated version of our model that uses only random negative candidates during training show that hard negative mining is essential for the model for better entity disambiguation.

\begin{figure}[h]
    \centering
    \includegraphics[width=0.8\linewidth]{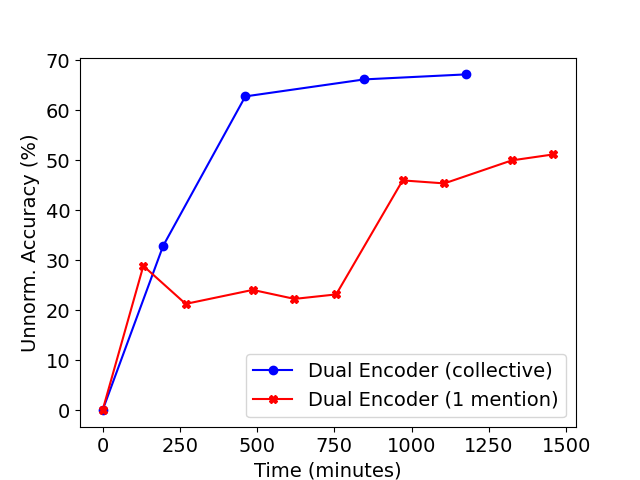}
    \caption{Comparative analysis of training speed measured in terms of accuracy achieved in first 24 hours of training. Both models were trained on 4 NVIDIA Quadro RTX GPUs with 24 GB memory.}
    \label{fig:accuracy}
\end{figure}

\begin{figure}[h!]
    \centering
    \includegraphics[width=0.8\linewidth]{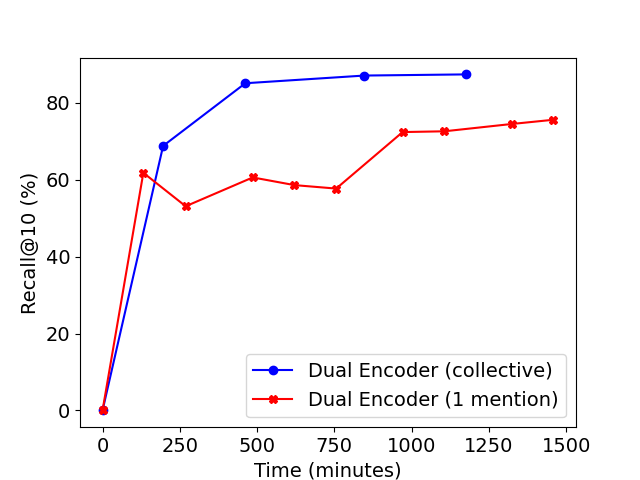}
    \caption{Comparative analysis of training speed measured in terms of recall@10 achieved in first 24 hours of training. Both models were trained on 4 NVIDIA Quadro RTX GPUs with 24 GB memory.}
    \label{fig:recall}
\end{figure}

\paragraph{Training and Inference Speed}
\label{training_inference_speed}
We perform a comparative analysis of the training speed of our collective dual encoder model with the single-mention dual encoder model. We show in Fig. \ref{fig:accuracy} and \ref{fig:recall} that our model achieves higher accuracy and recall@10 much faster than the single-mention dual encoder model. In fact, our model is 3x faster than the single-mention Dual Encoder model.

\begin{table}[h!]
    \centering
    \begin{tabular}{l|r}
    Model & mentions/sec\\
    \hline
        BLINK & 11.5\\
        Dual Encoder (1 mention) & 65.0\\
        Dual Encoder (collective) & \textbf{192.4}\\
    \hline
    \end{tabular}
    \caption{Inference speed comparison on the MedMentions dataset.}
    \label{tab:speed_medmentions}
\end{table}

\begin{table}[h!]
    \centering
    \begin{tabular}{l|r}
    Model & mentions/sec\\
    \hline
        BLINK & 11.5\\
        Dual Encoder (1 mention) & 87.0\\
        Dual Encoder (collective) & \textbf{402.5}\\
    \hline
    \end{tabular}
    \caption{Inference speed comparison on the BC5CDR dataset.}
    \label{tab:speed_bc5cdr}
\end{table}

We also compare the inference speed of our model with BLINK and the single-mention Dual Encoder model. The comparisons of inference speed for the two datasets are presented in Tables \ref{tab:speed_medmentions} and \ref{tab:speed_bc5cdr}, respectively. The inference speed is measured on a single NVIDIA Quadro RTX GPU with batch size 1. We observe that our collective dual encoder model is 3-4x faster than the single-mention Dual Encoder model and up to 25x faster (on average over the two datasets) than BLINK.  Since our model can process a document with $N$ mentions in one shot, we achieve higher entity disambiguation speed than the single-mention Dual Encoder and the BLINK model -- both require $N$ forward passes to process the $N$ mentions in a document. For these experiments, we set $N=8$, i.e., our collective dual encoder model processes up to $8$ mentions in a single pass. Note that the value of $N$ could be increased further for the inference phase. Caching the entity representations also helps our model and the single-mention Dual Encoder model at test time. The cross encoder of BLINK prevents it from using any cached entity representations, which drastically slows down the entity resolution speed of BLINK.

\paragraph{Candidate Recall}

\begin{table*}[htbp]
    \centering
    \begin{tabular}{l|c|c|c|c}
    & \multicolumn{2}{c|}{MedMentions} & \multicolumn{2}{c}{BC5CDR} \\
        \hline
        Model & Dev & Test & Dev & Test\\
        \hline
        BM25 & 59.8 & 59.5 & 76.3 & 74.5 \\
        Dense retrieval (1 mention) & 80.2 & 80.6 & 92.1 & \textbf{92.3}\\
        Dense retrieval (collective) & \textbf{87.5} & \textbf{87.6} & \textbf{92.2} & \textbf{92.3}\\
        \hline
    \end{tabular}
    \caption{Comparison of development and test set Recall@10 on MedMentions and BC5CDR datasets}
    \label{tab:recall_medmentions}
\end{table*}

We compare the recall@10 metrics of BM25 retrieval method used in LATTE and Cross Encoder to the dense retrieval method used in BLINK and in our model. We present our results in Tables \ref{tab:recall_medmentions}
for the MedMentions and BC5CDR datasets, respectively. Similar to the observations made for BLINK and  \newcite{gillick-etal-2019-learning}, we also find that dense retrieval has a superior recall than BM25. However, we observe that the recall value of dense retrieval depends on the underlying entity disambiguation model. For instance, on the MedMentions dataset, our model has much higher recall@10 than the Dual Encoder model that processes each mention independently, while both models are trained using a combination of hard and random negative candidates. However, this observation is not consistent across datasets as we do not observe similar gains in recall@10 for the BC5CDR dataset. We will explore this phenomenon in future work.

\paragraph{End-to-End Entity Disambiguation}

\begin{table*}[htbp]
\centering
\resizebox{16cm}{!}{
\begin{tabular}{l|c|c|c|c|c|c|c|c|c|c|c|c}
    & \multicolumn{6}{c|}{MedMentions} & \multicolumn{6}{c}{BC5CDR}\\
    \hline
    Model & \multicolumn{3}{c|}{Partial match} & \multicolumn{3}{c|}{Strict match} & \multicolumn{3}{c|}{Partial match} & \multicolumn{3}{c}{Strict match} \\
     & P & R & F1 & P & R & F1  & P & R & F1 & P & R & F1 \\
    \hline
    SciSpacy & 40.9 & 40.2 & 40.6 & 37.7 & 36.6 & 37.1 & 15.5 & 53.4 & 24.0 & 14.5 & 48.4 & 22.3 \\
    MedType & 44.7 & 44.1 & 44.4 & 41.2 & 40.0 & 40.6 & 16.6 & 57.0 & 25.7 & 15.3 & 51.0 & 23.5 \\ 
    \hline
    Dual Encoder (BIO tags) & 44.5 & 37.6 & 40.7 & 41.2 & 34.9 & 37.8 & 29.2 & 31.5 & 30.3 & 10.2 & 10.8 & 10.5 \\
    Dual Encoder (Exhaustive) & \textbf{56.3} & \textbf{56.4} & \textbf{56.4} & \textbf{52.9} & \textbf{53.8} & \textbf{53.4} & \textbf{76.0} & \textbf{74.4} & \textbf{75.2} & \textbf{74.6} & \textbf{73.1} & \textbf{73.8} \\
    \hline
\end{tabular}
}
\caption{Micro Precision (P), Recall (R) and F1 scores for the end-to-end entity linking task on the MedMentions and BC5DCR datasets.}
\label{tab:results_e2e}
\end{table*}

For the end-to-end entity linking task, we evaluate the models with two different evaluation protocols. In the \emph{strict match} protocol, the predicted mention spans and predicted target entity must match strictly with the gold spans and target entity. In the \emph{partial match} protocol, if there is an overlap between the predicted mention span and the gold mention span, and the predicted target entity matches the gold target entity, then it is considered to be a true positive. We evaluate our models using micro-averaged precision, recall, and F1 scores as evaluation metrics. For a fair comparison, we use the off-the-shelf evaluation tool  \texttt{neleval}\footnote{https://github.com/wikilinks/neleval}, which is also used for MedType. We follow the same evaluation protocol and settings as used for MedType.

We present the results of our collective Dual Encoder model and the baselines in Table \ref{tab:results_e2e}. The results show that exhaustive search over all possible spans for mention detection yields significantly better results than the \texttt{BIO} tagging based method, despite the additional computational cost. Moreover, our dual encoder based end-to-end entity linking model significantly outperforms SciSpacy and MedType. Note that there are highly specialized models such as TaggerOne \cite{leaman2016taggerone} that perform much better than our model on the BC5CDR dataset. However, TaggerOne is suitable for a few specific types of entities such as \emph{Disease} and \emph{Chemical}. For a dataset with entities of various different semantic types (e.g., MedMentions), \newcite{Mohan2019MedMentionsAL} show that TaggerOne performs inadequately. For such datasets where the target entities belong to many different semantic types, our proposed model is more effective as compared to highly specialized models like TaggerOne.

\section{Conclusion}
This paper introduces a biomedical entity linking approach using BERT-based dual encoders to disambiguate multiple mentions of biomedical concepts in a document in a single shot. We show empirically that our method achieves higher accuracy and recall than other competitive baseline models in significantly less training and inference time. We also showed that our method is significantly better than two recently proposed biomedical entity linking models for the end-to-end entity disambiguation task when subjected to multi-task learning objectives for joint mention span detection and entity disambiguation using a single model. 

\section*{Acknowledgments} We thank Vipina Kuttichi Keloth for her generous assistance in data processing and initial experiments. We thank Diffbot and the Google Cloud Platform for granting us access to computing infrastructure used to run some of the experiments reported in this paper.

\bibliography{eacl2021}

\begin{thebibliography}{22}
\expandafter\ifx\csname natexlab\endcsname\relax\def\natexlab#1{#1}\fi

\bibitem[{Aronson(2006)}]{aronson2006metamap}
Alan~R Aronson. 2006.
\newblock Metamap: Mapping text to the umls metathesaurus.
\newblock \emph{Bethesda, MD: NLM, NIH, DHHS}, 1:26.

\bibitem[{Beltagy et~al.(2020)Beltagy, Peters, and
  Cohan}]{Beltagy2020Longformer}
Iz~Beltagy, Matthew~E. Peters, and Arman Cohan. 2020.
\newblock Longformer: The long-document transformer.
\newblock \emph{arXiv:2004.05150}.

\bibitem[{Devlin et~al.(2019)Devlin, Chang, Lee, and
  Toutanova}]{devlin-etal-2019-bert}
Jacob Devlin, Ming-Wei Chang, Kenton Lee, and Kristina Toutanova. 2019.
\newblock \href {https://doi.org/10.18653/v1/N19-1423} {{BERT}: Pre-training of
  deep bidirectional transformers for language understanding}.
\newblock In \emph{Proceedings of the 2019 Conference of the North {A}merican
  Chapter of the Association for Computational Linguistics: Human Language
  Technologies, Volume 1 (Long and Short Papers)}, pages 4171--4186,
  Minneapolis, Minnesota. Association for Computational Linguistics.

\bibitem[{Furrer et~al.(2020)Furrer, Cornelius, and
  Rinaldi}]{furrer2020parallel}
Lenz Furrer, Joseph Cornelius, and Fabio Rinaldi. 2020.
\newblock \href {http://arxiv.org/abs/2003.07424} {Parallel sequence tagging
  for concept recognition}.

\bibitem[{Févry et~al.(2020)Févry, FitzGerald, Soares, and
  Kwiatkowski}]{fvry2020empirical}
Thibault Févry, Nicholas FitzGerald, Livio~Baldini Soares, and Tom
  Kwiatkowski. 2020.
\newblock \href {http://arxiv.org/abs/2005.14253} {Empirical evaluation of
  pretraining strategies for supervised entity linking}.

\bibitem[{Gillick et~al.(2019)Gillick, Kulkarni, Lansing, Presta, Baldridge,
  Ie, and Garcia-Olano}]{gillick-etal-2019-learning}
Daniel Gillick, Sayali Kulkarni, Larry Lansing, Alessandro Presta, Jason
  Baldridge, Eugene Ie, and Diego Garcia-Olano. 2019.
\newblock \href {https://doi.org/10.18653/v1/K19-1049} {Learning dense
  representations for entity retrieval}.
\newblock In \emph{Proceedings of the 23rd Conference on Computational Natural
  Language Learning (CoNLL)}, pages 528--537, Hong Kong, China. Association for
  Computational Linguistics.

\bibitem[{{Hogan} et~al.(2020){Hogan}, {Blomqvist}, {Cochez}, {d'Amato}, {de
  Melo}, {Gutierrez}, {Labra Gayo}, {Kirrane}, {Neumaier}, {Polleres},
  {Navigli}, {Ngonga Ngomo}, {Rashid}, {Rula}, {Schmelzeisen}, {Sequeda},
  {Staab}, and {Zimmermann}}]{KnowledgeGraphs2020}
Aidan {Hogan}, Eva {Blomqvist}, Michael {Cochez}, Claudia {d'Amato}, Gerard {de
  Melo}, Claudio {Gutierrez}, Jos{\'e}~Emilio {Labra Gayo}, Sabrina {Kirrane},
  Sebastian {Neumaier}, Axel {Polleres}, Roberto {Navigli}, Axel-Cyrille
  {Ngonga Ngomo}, Sabbir~M. {Rashid}, Anisa {Rula}, Lukas {Schmelzeisen}, Juan
  {Sequeda}, Steffen {Staab}, and Antoine {Zimmermann}. 2020.
\newblock \href {https://arxiv.org/abs/2003.02320} {Knowledge graphs}.
\newblock \emph{ArXiv}, 2003.02320.

\bibitem[{Ji et~al.(2020)Ji, Wei, and Xu}]{ji2020bert}
Zongcheng Ji, Qiang Wei, and Hua Xu. 2020.
\newblock Bert-based ranking for biomedical entity normalization.
\newblock \emph{AMIA Summits on Translational Science Proceedings}, 2020:269.

\bibitem[{Kolitsas et~al.(2018)Kolitsas, Ganea, and
  Hofmann}]{kolitsas-etal-2018-end}
Nikolaos Kolitsas, Octavian-Eugen Ganea, and Thomas Hofmann. 2018.
\newblock \href {https://doi.org/10.18653/v1/K18-1050} {End-to-end neural
  entity linking}.
\newblock In \emph{Proceedings of the 22nd Conference on Computational Natural
  Language Learning}, pages 519--529, Brussels, Belgium. Association for
  Computational Linguistics.

\bibitem[{Leaman and Lu(2016)}]{leaman2016taggerone}
Robert Leaman and Zhiyong Lu. 2016.
\newblock Taggerone: joint named entity recognition and normalization with
  semi-markov models.
\newblock \emph{Bioinformatics}, 32(18):2839--2846.

\bibitem[{Lee et~al.(2019)Lee, Yoon, Kim, Kim, Kim, So, and
  Kang}]{BioBERT:Lee_2019}
Jinhyuk Lee, Wonjin Yoon, Sungdong Kim, Donghyeon Kim, Sunkyu Kim, Chan~Ho So,
  and Jaewoo Kang. 2019.
\newblock \href {https://doi.org/10.1093/bioinformatics/btz682} {Biobert: a
  pre-trained biomedical language representation model for biomedical text
  mining}.
\newblock \emph{Bioinformatics}.

\bibitem[{Li et~al.(2020)Li, Min, Iyer, Mehdad, and
  Yih}]{li-etal-2020-efficient}
Belinda~Z. Li, Sewon Min, Srinivasan Iyer, Yashar Mehdad, and Wen-tau Yih.
  2020.
\newblock \href {https://doi.org/10.18653/v1/2020.emnlp-main.522} {Efficient
  one-pass end-to-end entity linking for questions}.
\newblock In \emph{Proceedings of the 2020 Conference on Empirical Methods in
  Natural Language Processing (EMNLP)}, pages 6433--6441, Online. Association
  for Computational Linguistics.

\bibitem[{Li et~al.(2016)Li, Sun, Johnson, Sciaky, Wei, Leaman, Davis,
  Mattingly, Wiegers, and Lu}]{li2016biocreative}
Jiao Li, Yueping Sun, Robin~J Johnson, Daniela Sciaky, Chih-Hsuan Wei, Robert
  Leaman, Allan~Peter Davis, Carolyn~J Mattingly, Thomas~C Wiegers, and Zhiyong
  Lu. 2016.
\newblock Biocreative v cdr task corpus: a resource for chemical disease
  relation extraction.
\newblock \emph{Database}, 2016.

\bibitem[{Logeswaran et~al.(2019)Logeswaran, Chang, Lee, Toutanova, Devlin, and
  Lee}]{logeswaran-etal-2019-zero}
Lajanugen Logeswaran, Ming-Wei Chang, Kenton Lee, Kristina Toutanova, Jacob
  Devlin, and Honglak Lee. 2019.
\newblock \href {https://doi.org/10.18653/v1/P19-1335} {Zero-shot entity
  linking by reading entity descriptions}.
\newblock In \emph{Proceedings of the 57th Annual Meeting of the Association
  for Computational Linguistics}, pages 3449--3460, Florence, Italy.
  Association for Computational Linguistics.

\bibitem[{Mohan and Li(2019)}]{Mohan2019MedMentionsAL}
Sunil Mohan and Donghui Li. 2019.
\newblock Medmentions: A large biomedical corpus annotated with umls concepts.
\newblock \emph{ArXiv}, abs/1902.09476.

\bibitem[{Neumann et~al.(2019)Neumann, King, Beltagy, and
  Ammar}]{neumann-etal-2019-scispacy}
Mark Neumann, Daniel King, Iz~Beltagy, and Waleed Ammar. 2019.
\newblock \href {https://doi.org/10.18653/v1/W19-5034} {{S}cispa{C}y: {F}ast
  and {R}obust {M}odels for {B}iomedical {N}atural {L}anguage {P}rocessing}.
\newblock In \emph{Proceedings of the 18th BioNLP Workshop and Shared Task},
  pages 319--327, Florence, Italy. Association for Computational Linguistics.

\bibitem[{Savova et~al.(2010)Savova, Masanz, Ogren, Zheng, Sohn,
  Kipper-Schuler, and Chute}]{savova2010ctakes}
Guergana~K Savova, James~J Masanz, Philip~V Ogren, Jiaping Zheng, Sunghwan
  Sohn, Karin~C Kipper-Schuler, and Christopher~G Chute. 2010.
\newblock Mayo clinical text analysis and knowledge extraction system (ctakes):
  architecture, component evaluation and applications.
\newblock \emph{Journal of the American Medical Informatics Association},
  17(5):507--513.

\bibitem[{Soldaini and Goharian(2016)}]{soldaini2016quickumls}
Luca Soldaini and Nazli Goharian. 2016.
\newblock Quickumls: a fast, unsupervised approach for medical concept
  extraction.
\newblock In \emph{MedIR workshop, sigir}, pages 1--4.

\bibitem[{{Vashishth} et~al.(2020){Vashishth}, {Joshi}, {Dutt},
  {Newman-Griffis}, and {Rose}}]{medtype2020}
Shikhar {Vashishth}, Rishabh {Joshi}, Ritam {Dutt}, Denis {Newman-Griffis}, and
  Carolyn {Rose}. 2020.
\newblock \href {http://arxiv.org/abs/2005.00460} {{MedType: Improving Medical
  Entity Linking with Semantic Type Prediction}}.
\newblock \emph{arXiv e-prints}, page arXiv:2005.00460.

\bibitem[{Wu et~al.(2019)Wu, Petroni, Josifoski, Riedel, and
  Zettlemoyer}]{BLINK:2019:Wu}
Ledell Wu, Fabio Petroni, Martin Josifoski, Sebastian Riedel, and Luke
  Zettlemoyer. 2019.
\newblock Zero-shot entity linking with dense entity retrieval.
\newblock In \emph{arXiv:1911.03814}.

\bibitem[{Xu et~al.(2020)Xu, Zhang, and Bethard}]{xu-etal-2020-generate}
Dongfang Xu, Zeyu Zhang, and Steven Bethard. 2020.
\newblock \href {https://doi.org/10.18653/v1/2020.acl-main.748} {A
  generate-and-rank framework with semantic type regularization for biomedical
  concept normalization}.
\newblock In \emph{Proceedings of the 58th Annual Meeting of the Association
  for Computational Linguistics}, pages 8452--8464, Online. Association for
  Computational Linguistics.

\bibitem[{Zhu et~al.(2019)Zhu, Celikkaya, Bhatia, and Reddy}]{LATTE:Zhu:2019}
Ming Zhu, Busra Celikkaya, Parminder Bhatia, and Chandan~K. Reddy. 2019.
\newblock \href {http://arxiv.org/abs/1911.09787} {Latte: Latent type modeling
  for biomedical entity linking}.

\end{thebibliography}
\bibliographystyle{acl_natbib}
\end{document}